\documentclass[pdflatex,sn-mathphys-num]{sn-jnl}

\usepackage{graphicx}%
\usepackage{multirow}%
\usepackage{amsmath,amssymb,amsfonts}%
\usepackage{amsthm}%
\usepackage{mathrsfs}%
\usepackage[title]{appendix}%
\usepackage{xcolor}%
\usepackage{textcomp}%
\usepackage{manyfoot}%
\usepackage{booktabs}%
\usepackage{algorithm}%
\usepackage{algorithmicx}%
\usepackage{algpseudocode}%
\usepackage{listings}%

\usepackage{arydshln}

\usepackage{tabularx}

\theoremstyle{thmstyleone}%
\theoremstyle{thmstyletwo}%

\theoremstyle{thmstylethree}%

\begin{document}

\title[Multi-Intent Spoken Language Understanding: Methods, Trends, and Challenges]{Multi-Intent Spoken Language Understanding: Methods, Trends, and Challenges}


\author[1,2]{\fnm{Di} \sur{Wu}} \equalcont{These authors contributed equally to this work.}

\author[2]{\fnm{Ruiyu} \sur{Fang}} \equalcont{These authors contributed equally to this work.}

\author[1]{\fnm{Liting} \sur{Jiang}} \equalcont{These authors contributed equally to this work.}

\author[2]{\fnm{Shuangyong} \sur{Song}} \equalcont{These authors contributed equally to this work.}

\author[2]{\fnm{Xiaomeng} \sur{Huang}}

\author[2]{\fnm{Shiquan} \sur{Wang}}

\author[2]{\fnm{Zhongqiu} \sur{Li}}

\author[2]{\fnm{Lingling} \sur{Shi}}

\author[2]{\fnm{Mengjiao} \sur{Bao}}

\author*[2]{\fnm{Yongxiang} \sur{Li}}\email{liyx25@chinatelecom.cn}


\author*[1,3]{\fnm{Hao} \sur{Huang}}\email{huanghao@xju.edu.cn}

\affil[1]{School of Computer Science and Technology, Xinjiang University}

\affil[2]{Institute of Artificial Intelligence (TeleAI), China Telecom Corp Ltd}

\affil[3]{Joint International Research Laboratory of Silk Road Multilingual Cognitive Computing, Urumqi, China}


\abstract{Multi-intent spoken language understanding (SLU) involves two tasks: multiple intent detection and slot filling, which jointly handle utterances containing more than one intent. Owing to this characteristic, which closely reflects real-world applications, the task has attracted increasing research attention, and substantial progress has been achieved. However, there remains a lack of a comprehensive and systematic review of existing studies on multi-intent SLU. To this end, this paper presents a survey of recent advances in multi-intent SLU. We provide an in-depth overview of previous research from two perspectives: decoding paradigms and modeling approaches. On this basis, we further compare the performance of representative models and analyze their strengths and limitations. Finally, we discuss the current challenges and outline promising directions for future research. We hope this survey will offer valuable insights and serve as a useful reference for advancing research in multi-intent SLU.

}




\maketitle

\section{Introduction}\label{sec1}

\begin{figure}[h]
\centering
\includegraphics[]{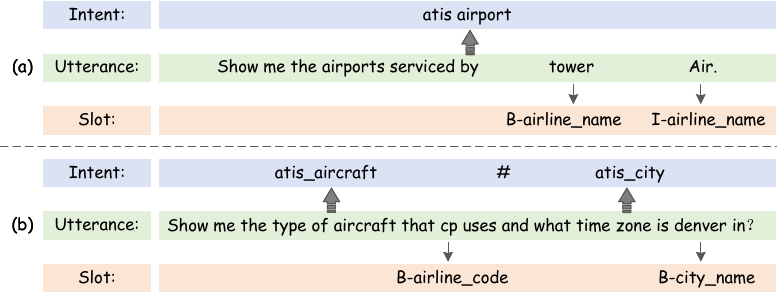}
\caption{Examples from the SLU dataset. (a) shows a single-intent SLU sample, while (b) presents a multi-intent SLU sample. For clarity, slot labels marked as ``O” are omitted.}\label{fig_intro}
\end{figure}

Spoken Language Understanding (SLU) typically involves two core tasks: intent detection and slot filling \cite{tur2011spoken}. The former identifies the request underlying goal, while the latter extracts entities related to that goal from the utterance. As a key component of task-oriented dialogue systems, SLU enables systems to interpret user input and respond appropriately. As shown in Fig. \ref{fig_intro}(a), given the user utterance ``Show me the airports serviced by tower Air”, the intent is ``atis\_airport”, and the corresponding slot annotations are ``O O O O O O B-airline\_name I-airline\_name”.

In real-world scenarios, a single utterance may express multiple intents, about 52\% of utterances in Amazon's internal dataset contain multiple intents \cite{gangadharaiah2019joint}. For instance, as shown in Fig.~\ref{fig_intro}(b), the user utterance ``Show me the type of aircraft that cp uses and what time zone is denver in?'' conveys multiple intents (``atis\_aircraft\#atis\_city'') with corresponding slot annotations ``O O O O O O O B-airline\_code O O O O O O B-city\_name O''.

Although multi-intent SLU shares a similar objective with single-intent SLU, namely predicting intents and slot labels from utterances, it poses unique challenges. Multi-intent utterances often comprise multiple clauses, making it difficult for single-intent models to capture all semantic dependencies \cite{liu2016attention,goo2018slot,wang2018bi,haihong2019novel,qin2019stack}. The challenges mainly arise from two aspects:
1. \textbf{Complex decoding mechanisms}. Single-intent SLU models usually encode an utterance into a single sentence-level representation for intent prediction, which limits their ability to distinguish multiple co-occurring intents.
2. \textbf{Increased interaction complexity between intent and slot features}. As illustrated in Fig.~\ref{fig_intro}, intent detection and slot filling are interdependent. However, in multi-intent utterances, cross-clause interference exacerbates the difficulty of modeling this interaction effectively.

\begin{figure}[h]
\centering
\includegraphics[]{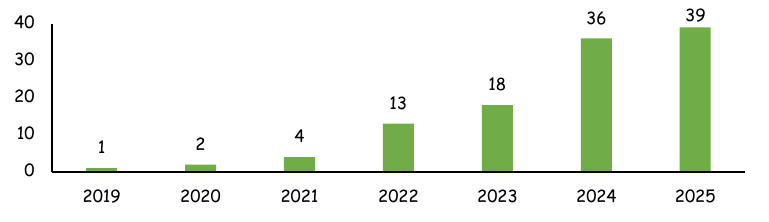}
\caption{Trend in the cumulative number of multi-intent SLU papers.}\label{1}
\end{figure}

Recognizing the importance of multi-intent SLU for building more capable dialogue systems, researchers have made significant progress in recent years. The number of related publications has grown steadily (Fig.~\ref{1}). Nevertheless, existing surveys only briefly mention or define multi-intent SLU without providing a comprehensive and systematic review \cite{qin2021survey,weld2022survey,louvan2020recent}.

To this end, we present the first in-depth survey of multi-intent SLU research since 2019. We systematically analyze representative decoding and modeling methods, summarize how they establish information flow between multiple intent detection and slot filling, and examine how design choices influence model performance. We also discuss current limitations and outline promising directions for future research.

The main contributions of this paper are as follows:
\begin{itemize}

\item We conduct a systematic comparison and analysis of existing multi-intent SLU studies from two perspectives: decoding methods (classification-based, and generation-based) and modeling strategies (intent-guided slot filling, slot-guided intent detection, and bidirectional guidance).

\item We summarize the strengths and weaknesses of different approaches and provide links to publicly available implementations to facilitate future research.

\item We discuss the current limitations of multi-intent SLU and highlight potential research directions and open challenges in this emerging field.

\end{itemize}

The remainder of this paper is organized as follows. $\S$ \ref{sec_related} reviews prior SLU survey studies and highlights the distinctions between their focus and ours. $\S$ \ref{sec2} introduces the fundamental concepts of multi-intent SLU, including task definitions, datasets, evaluation metrics. $\S$ \ref{sec2} introduces the fundamental concepts of multi-intent SLU, including task definitions, datasets, evaluation metrics. 
$\S$ \ref{sec_de} details several methods to decode intents and slots.
$\S$ \ref{sec3} presents a detailed review of existing modeling approaches, analyzes their characteristics, and discusses how different strategies impact model performance. $\S$ \ref{sec4} examines unresolved challenges and potential research directions from three perspectives: datasets, methodologies, and practical applications. Finally, $\S$ \ref{sec5} concludes the paper.

\section{Related work}\label{sec_related}
In recent years, joint modeling of intent detection and slot filling has emerged as one of the central research directions in task-oriented dialogue systems, with an increasing number of survey studies contributing to this area.
\citet{louvan2020recent} categorized existing methods based on model architectures into independent models, joint models, and transfer learning models, emphasizing the performance of neural approaches in single-domain and single-turn scenarios.
\citet{weld2022survey} focused specifically on joint models, systematically reviewing the architectural evolution from implicit sharing to explicit interaction and providing the first detailed comparison between unidirectional and bidirectional information flows.
\citet{qin2021survey} proposed a new taxonomy that summarizes the development of SLU techniques across three paradigms, namely single-task, joint, and pre-trained models (PTMs), while also addressing emerging challenges such as multi-turn, multi-intent, cross-domain, and cross-lingual scenarios. 
More recently, \citet{muhammad2025joint}, offering the survey with the broadest historical coverage, traced the evolution of joint learning methods from classical approaches such as Conditional Random Fields (CRF) \cite{sutton2012introduction} and Support Vector Machines (SVM) \cite{hearst1998support} to PTMs like Transformer \cite{vaswani2017attention} and RoBERTa \cite{liu2019roberta}. They introduced a threefold classification of architectures, namely implicit, explicit, and hybrid, and provided an in-depth analysis of performance differences and dataset-related challenges.

Although \citet{weld2022survey, qin2021survey, muhammad2025joint} discuss multi-intent SLU and acknowledge that it is more challenging and closer to real-world scenarios than traditional single-intent SLU, none of them offer a comprehensive and systematic overview of the field.
In contrast, this survey focuses exclusively on multi-intent SLU, reviewing the differences and characteristics of various intent and slot decoding strategies, and categorizing existing methods according to their patterns of information interaction. Furthermore, we analyze the limitations of current approaches, identify key research gaps, and highlight potential directions for future work. We hope that this survey will serve as a valuable resource for researchers and practitioners, facilitating a deeper understanding of multi-intent SLU and inspiring future advancements in this area.

\section{Task Description}\label{sec2}
In this section, we introduce the task of multi-intent SLU, including the definitions of multiple intent detection and slot filling ($\S$ \ref{sec21}), as well as commonly used multi-intent SLU datasets ($\S$ \ref{sec22}) and evaluation metrics ($\S$ \ref{sec23}).

\subsection{Task Definition}\label{sec21}
In multi-intent SLU, given a user utterance $U=(u_1, u_2, \ldots ,u_n)$ as input, the model aims to predict the intents $O^{I}$ and slots $O^{S}$ expressed in the utterance. Where, $n$ denotes the length of the utterance.
\begin{equation}
O^{I}, O^{S} = Model(U).
\end{equation}

\subsection{Datasets}\label{sec22}
Two commonly used datasets for multi-intent SLU are MixATIS and MixSNIPS \cite{qin2020agif}. MixATIS is derived from the ATIS dataset \cite{hemphill1990atis}, where utterances corresponding to different intents are concatenated with the conjunction ``and”. The proportions of single-, double-, and triple-intent utterances are approximately [0.3, 0.5, 0.2]. MixSNIPS is constructed in a similar manner based on the SNIPS dataset \cite{coucke2018snips}. The statistics of both datasets are summarized in Table~\ref{tab:dataset_stats}.
\begin{table}[h]
\renewcommand\arraystretch{1}
\centering
\caption{Statistics of MixATIS and MixSNIPS datasets.}
\label{tab:dataset_stats}
\begin{tabular}{lccccc}
\toprule
Dataset & Training & Validation & Test & Intent & Slot\\
\midrule
MixATIS & 13,162 & 756 & 828 & 18 & 117\\
MixSNIPS & 39,776 & 2,198 & 2,199 & 7 & 72\\
\bottomrule
\end{tabular}
\end{table}

\subsection{Evaluation Metrics}\label{sec23}
Consistent with previous studies on single-intent SLU \cite{goo2018slot, wu2024cea}, the commonly used evaluation metrics for multi-intent SLU include intent accuracy (Acc), slot F1, and overall accuracy (Acc).
\begin{itemize}
\item \textbf{Intent Acc}: Evaluates the performance of multiple intent detection. A prediction is regarded as correct only when all intents in the utterance are correctly identified.

\item \textbf{Slot F1}: Measures the performance of slot recognition. It is defined as the harmonic mean of precision and recall, where a slot prediction is considered correct only if it exactly matches the ground truth.

\item \textbf{Overall Acc}: Reflects the proportion of utterances for which both intents and slots are correctly predicted. This metric jointly accounts for the performance of multiple intent detection and slot filling.

\end{itemize}

\section{Decoding Method}\label{sec_de}
Unlike single-intent SLU, utterances in multi-intent SLU datasets may contain multiple intents, and their decoding methods differ accordingly. In this section, we review the different decoding approaches for intents ($\S$ \ref{sec_intent}) and slots ($\S$ \ref{sec_slot}).

\subsection{Multiple Intent Decoding Method}\label{sec_intent}
In previous single-intent SLU studies \cite{qin2021co,hao2023joint}, intent detection is often treated as a text classification problem \cite{minaee2021deep,pang2022mfdg,xu2023improving,xiong2024dual,zhao2022tosa,zhang2022multi,song2015classifying}. The key difference between multiple intent detection and intent detection lies in the number of intents contained in an utterance $U$, as a user may express multiple intents in a single sentence. Therefore, the decoding methods for multiple intent detection differ significantly from those for intent detection.
The commonly used intent decoding approaches in prior multi-intent SLU research include the following:

\subsubsection{Threshold-based Multi-label Classification Approach}
\begin{figure}[h]
\centering
\includegraphics[]{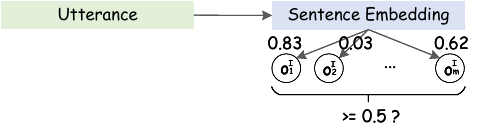}
\caption{Threshold-based Multi-label classification Approach.}\label{fig_intent_decoding_1}
\end{figure}

As shown in Fig. \ref{fig_intent_decoding_1}, \citet{qin2020agif} defined multiple intent detection as a multi-label text classification problem \cite{ma2021label}. Specifically, the entire user utterance is encoded into a sentence vector, which is then used to score each candidate intent label. A sigmoid function is applied to produce probabilities between 0 and 1, and a threshold is set to determine which intents are present in the utterance. For example, when the threshold is set to 0.5 and the output probabilities are $\left [  0.83, 0.03, \ldots ,0.62 \right ]$, the intent labels whose predicted probabilities exceed 0.5 are identified as the detected intents.

\noindent \textbf{Advantages}: The approach of directly performing multi-label classification using sentence embeddings is straightforward and easy to implement.

\noindent \textbf{Limitations}: This method is sensitive to the choice of threshold, and all tokens share the same intent representation, failing to distinguish between tokens from different sub-utterances or phrases.

\subsubsection{Threshold-based Token Voting Approach}
\begin{figure}[h]
\centering
\includegraphics[]{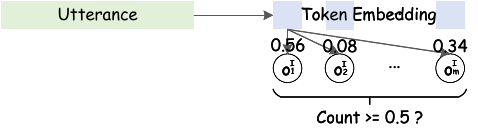}
\caption{Threshold-based Token Voting Approach.}\label{fig_intent_decoding_2}
\end{figure}

As shown in Fig. \ref{fig_intent_decoding_2}, \citet{qin2021gl} proposed a method in which each token in the utterance participates in voting. After processing through a sigmoid layer, each token receives a value between 0 and 1. The votes for each intent are aggregated by summing over all tokens. An intent is included in the final prediction if and only if the total number of votes it obtains exceeds half of the utterance's length. For instance, for an utterance of length 7, the threshold is automatically set to 4. If the vote counts across intents are $\left [6, 5, 1, 1,2 \right ]$, only those receiving $\geq$ 4 votes (i.e., 6 and 5) are selected, and the corresponding intent labels are output.

\noindent\textbf{Advantages}: Each token participates in the intent decoding process. Compared with compressing the utterance into a single sentence embedding, this approach is more likely to achieve higher accuracy.

\noindent\textbf{Limitations}: Although the method appears to eliminate the need for a manually set threshold, it essentially uses the utterance length as the threshold. This can still be problematic when a sub-clause within the utterance is very short.

\subsubsection{Threshold-free Token Voting Approach}
\begin{figure}[h]
\centering
\includegraphics[]{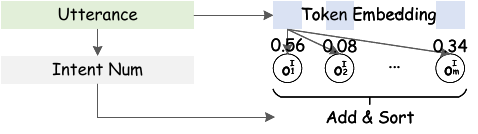}
\caption{Threshold-free Token Voting Approach.}\label{fig_intent_decoding_3}
\end{figure}

\citet{chen2022transformer} proposed a threshold-free voting method, as illustrated in Fig. \ref{fig_intent_decoding_3}. This approach introduces an additional subtask: intent number prediction, which aims to identify the number of intents $k$ present in the utterance. The probabilities of all token-level intent predictions are aggregated and sorted, and the top-k intents with the highest probabilities are selected as the final predicted intent labels. For example, if the intent count prediction subtask determines that k = 2, the summed intent probabilities across all tokens—say $\left [0.82, 0.75, 0.43, 0.38, 0.21\right ]$—are ranked, and the top two intent labels are chosen as output, eliminating the need for a sentence-length–based threshold.

\noindent\textbf{Advantages}: The method is not constrained by utterance length, allowing flexible prediction of any number of intents in both long and short utterances. Moreover, selecting the top-k intents by ranking naturally supports scenarios with single intent without requiring additional fallback mechanisms.

\noindent\textbf{Limitations}: The introduction of an additional subtask increases model complexity, necessitates the construction of intent count labels during training, and raises the risk of error propagation. A misprediction in the value of k (even by one) may lead to a complete mismatch in the final set of intents, making the approach highly sensitive to the accuracy of intent count estimation.

\subsubsection{Intent Label Generation Approach}
\begin{figure}[h]
\centering
\includegraphics[]{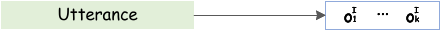}
\caption{Intent Label Generation Approach.}\label{fig_intent_decoding_4}
\end{figure}

A classic approach for generating labels involves using prompts that combine the user's utterance with a request for intent prediction, inputting them into a generative model to directly generate a sequence of intent labels in one step \cite{ning2023ump} (Fig. \ref{fig_intent_decoding_4}). This eliminates the need for thresholds, voting, or ranking mechanisms.

\noindent\textbf{Advantages}: Compared to carefully designed decoding methods, this direct generative approach is relatively simple and requires lower development effort.

\noindent\textbf{Limitations}: It is prone to hallucinations or formatting errors due to prompt sensitivity, suffers from slow autoregressive inference with high computational costs, and makes it difficult to trace the root causes of bad cases.


None of the aforementioned four approaches demonstrates a clear superiority, as each possesses its own advantages and limitations. The final multiple intent detection performance of a model is influenced not only by the choice of decoding strategy but also by factors such as utterance encoding methods and the approach to modeling interactions between intent and slot features. 


\subsection{Slot Decoding Method}\label{sec_slot}
In multi-intent SLU, slot filling is mainly approached in two ways: sequence labeling ($\S$ \ref{sec_intent}) and slot-value (span) pair generation ($\S$ \ref{sec_slot}).



\subsubsection{Sequence-labeling Approach}
\begin{figure}[h]
\centering
\includegraphics[]{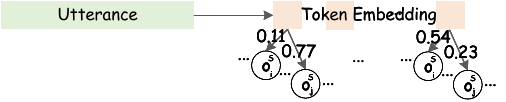}
\caption{Sequence-labeling Approach.}\label{fig_slot_decoding_1}
\end{figure}

As shown in Fig. \ref{fig_slot_decoding_1}, in this approach, each token in the utterance is assigned a label that indicates its semantic role or slot type. Common labeling schemes include the BIO (Begin, Inside, Outside) format and its variants \cite{lin2021asrnn}. The labels are then combined to form complete slot-value pairs, allowing the identification of entities and their corresponding roles within the utterance.

\noindent\textbf{Advantages}: This method is widely used due to its simplicity, interpretability, and effectiveness in capturing token-level slot information.

\noindent\textbf{Limitations}: Sequence labeling can be cumbersome when handling complex or nested semantic relations, making the annotation process tedious and increasing the likelihood of model errors.

\subsubsection{Slot-value Generation Approach}
\begin{figure}[h]
\centering
\includegraphics[]{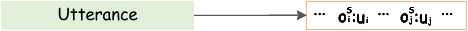}
\caption{Slot-value Generation Approach.}\label{fig_slot_decoding_2}
\end{figure}

As shown in Fig. \ref{fig_slot_decoding_2}, in this approach, the output is typically formatted as \texttt{<Value> is a <Slot>}, explicitly linking each extracted value to its corresponding slot type \cite{zhu2024zero,qin2025croprompt}. This method is commonly used in generative setups, where the system produces slot-value pairs directly as text. It can naturally accommodate varying slot vocabularies without requiring a fixed label set.

\noindent\textbf{Advantages}: Slot-value generation focuses on extracting global slot–value pairs, outputs key–value pairs, and is well suited for long sentences or nested structures, enabling efficient extraction of multiple slots.

\noindent\textbf{Limitations}: This approach demands strong semantic understanding; if the slot wording in the input differs from the expected phrasing, slot information may be lost or mispredicted.


In multi-intent SLU, sequence labeling is commonly used in classification-based small models or in approaches that employ PTMs for slot recognition. In contrast, slot-value generation is typically adopted in generation-based approach, where slots are often produced together with the intent. 
Furthermore, span prediction is also an important method for identifying entities in an utterance \cite{shen2023diffusionner,ding2024improving,fu2021spanner,wu2025int}. However, we found that this approach is not commonly used in multi-intent SLU, so we do not provide a detailed description here.

\section{Classification and Analysis of Modeling Approaches}\label{sec3}
\begin{figure}[h]
\centering
\includegraphics[]{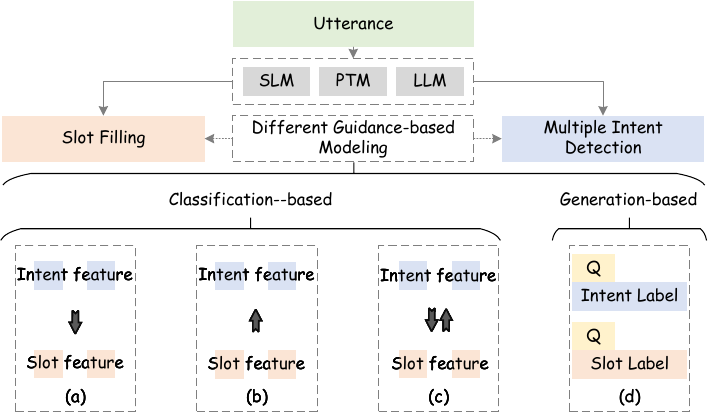}
\caption{Different Guidance Modeling Approaches for Multi-Intent SLU. Among them, (a) represents intent-to-slot guidance, (b) represents the reverse direction, (c) denotes mutual guidance between the two features, and (d) refers to instruction-based implicit guidance.}\label{interaction}
\end{figure}

In this section, we review previous work on multi-intent SLU. Rather than categorizing prior studies by the type of model employed, such as small language models (SLMs), PTMs, or large language models (LLMs), we classify and summarize them based on the guidance-oriented modeling approaches that connect the two tasks of multiple intent detection and slot filling, Fig. \ref{interaction}. The label prediction approaches can be divided into two categories: classification-based ($\S$ \ref{sec31}) and generation-based ($\S$ \ref{sec35}). We also review the performance of these approaches in $\S$ \ref{sec_res} and provided a corresponding analysis.

\subsection{Classification-based Approaches}\label{sec31}
The classification-based approaches include three guidance strategies: multiple intent guided slot filling ($\S$ \ref{sec32}), slot guided multiple intent detection ($\S$ \ref{sec33}), bidirectional interaction modeling ($\S$ \ref{sec34}). 

\subsubsection{Multiple Intent Guided Slot Filling}\label{sec32}
Multiple intent detection and slot filling are two closely related tasks, and establishing effective information transfer between them can substantially enhance model overall performance. Early studies primarily adopted the strategy of using multiple intents to guide slot filling (Fig. \ref{interaction} (a)). \citet{gangadharaiah2019joint} introduced a slot-gated mechanism that utilizes intent prediction results to guide slot filling. \citet{qin2020agif} proposed Adaptive Graph-Interactive Framework (AGIF), which employs a graph neural network to propagate information from intent nodes to slot nodes. Later, \citet{ding2021focus} refined this framework by incorporating an attention mechanism and proposed Dynamic Graph Model (DGM). Global-Locally Graph Interaction Network (GLGIN) \cite{qin2021gl} further improved upon AGIF by adopting a non-autoregressive decoding strategy, which significantly accelerates slot decoding. Building on GLGIN, \citet{gong2024mag} integrated the Mamba architecture to filter utterance features, thereby enhancing the model’s capability in intent and slot identification. 
\citet{huang2022clid} proposed Chunk Level Intent Detection (CLID), which focuses on segmenting intent clauses and uses the recognized intents to guide slot filling.
Subsequently, \citet{yin2024uni} further expanded the levels of interaction by establishing intent-to-slot guidance at deeper levels, including utterance-level, chunk-level, and token-level.
\citet{chen2022transformer} encoded utterances using a Transformer encoder and explicitly concatenated multiple intent detection results with slot features to guide slot filling. Similarly, \citet{zhu2024elsf} adopted a comparable information flow while introducing two additional tasks: slot boundary prediction and slot-entity type assignment.

\noindent\textbf{Advantages}: This modeling paradigm explicitly controls the information flow from multiple intent representation to slot filling, enabling the model to better exploit multiple intent knowledge for more accurate slot label prediction.

\noindent\textbf{Limitations}: The unidirectional information transfer may lead to information leakage from slot filling back to multiple intent detection, thereby constraining the overall model performance.

\subsubsection{Slot-Guided Multiple Intent Detection}\label{sec33}
Establishing the information flow from slots to multiple intent detection can likewise enhance model performance in multi-intent SLU tasks (Fig. \ref{interaction} (b)). \citet{cai2022slim} proposed Slot-Intent Mapping Model (SLIM), which introduces an explicit slot-intent classifier to model the many-to-one correspondence between slots and intents.

\noindent\textbf{Advantages}: Similar to $\S$ \ref{sec32}, this approach regulates the information flow from slots to multiple intents, effectively leveraging slot information to enable more accurate intent identification.

\noindent\textbf{Limitations}: This modeling strategy, however, remains unidirectional and is susceptible to the same issue of knowledge leakage.

\subsubsection{Bidirectional Interaction Modeling}\label{sec34}
The bidirectional interactive modeling approach has been widely adopted in multi-intent SLU tasks. This approach aims to comprehensively explore the correlations between multiple intent detection and slot filling by leveraging the knowledge from one task to assist the other (Fig. \ref{interaction} (c)). 

\citet{chen2022joint} proposed the Self-Distillation Joint Neural Language Understanding model (SDJN), which establishes a cyclic flow of information through three sequentially connected decoders. Building on the distillation framework, \citet{tu2023joint} further incorporated contrastive learning to enhance the model’s awareness of intent and slot label semantics. \citet{huang2024joint} employed a Graph Neural Network to capture the knowledge interactions between intent and slot representations. \citet{he2024conceptual} introduced the Conceptual Knowledge Enhanced Model (CKEM), which integrates ConceptNet \cite{speer2017conceptnet} as an external knowledge source alongside a graph network to improve intent and slot identification. \citet{sun2024position} enhanced information exchange between multiple intent detection and slot filling by exploiting token-level positional relationships, thereby strengthening the model’s understanding of inter-task dependencies.

The co-occurrence matrix between intent and slot labels has also proven effective for inter-task information transfer \cite{song2022enhancing}. \citet{wu2024dual} refined this mechanism by establishing information flow between intent labels and slot types and further proposed a Gaussian Graph Attention Network that leverages relative positional information to enhance semantic interactions. \citet{li2024hybrid} introduced the Graph Convolutional Network and Multi-head Transformer (GCN-MT) model, which considers slot label co-occurrence and incorporates both corpus-level and utterance-level contextual information to achieve cross-task knowledge transfer. Based on co-occurrence statistical dependencies, \citet{xing2022group} utilized label correlations to construct a topological graph and proposed the Recurrent Heterogeneous Label Matching Network (ReLa-Net) to improve intent and slot decoding through label dependency modeling.

\citet{pham2023misca} designed a joint attention mechanism between intent and slot labels, together with a basic-level label attention mechanism, to achieve task interaction without relying on complex graph structures. \citet{xing2022co} proposed a pre-recognition and mutual-guidance framework that eliminates dependence on external co-occurrence knowledge, providing a simpler yet effective alternative. The Scope-Sensitive Result Attention Network (SSRAN) focuses on intra-clause information propagation within the same intent segment of an utterance \cite{cheng2023scope}, thereby reducing cross-clause interference without requiring external knowledge. \citet{cheng2023accelerating} introduced a pluggable distillation mechanism that significantly accelerates inference for models such as ReLa-Net, Co-Guiding Net, and SSRAN, while maintaining competitive accuracy.

\citet{zhu2024aligner2} proposed dual alignment strategies—label prediction alignment and forced prediction alignment—which can be integrated into various multi-intent SLU models to enhance their individual performance. \citet{zhuang2024towards} defined cross-task information gain, which extracts informative inter-task knowledge through multi-stage iteration and multi-level cross-task contrastive learning. Finally, \citet{zhu2024dance} constructed a hybrid architecture combining a global static heterogeneous graph with a local dynamic heterogeneous interaction graph, enabling the model to differentiate intent and slot nodes and capture more complex inter-task relationships, thus improving its discriminative capability.

\noindent\textbf{Advantages}: 
This approach facilitates bidirectional information exchange between multiple intent detection and slot filling, effectively avoiding the knowledge leakage problem that may occur under unidirectional information flow. Compared with unidirectional guidance, bidirectional interaction enables the model to identify intents and slot labels with higher accuracy.

\noindent\textbf{Limitations}: 
Constructing a bidirectional guidance model slightly increases the number of parameters and computational cost relative to unidirectional models. In addition, bidirectional guidance may amplify error propagation, potentially leading to performance degradation.

\subsection{Generation-based Approaches}\label{sec35}
Many generation-based methods do not explicitly implement inter-task representation transfer; however, they can achieve information transfer between the multiple intent detection and slot filling tasks through unified instruction-based modeling.
Generative modeling for multi-intent SLU has emerged as an effective approach to mitigating the discrepancy between intent detection and slot filling. Rather than treating the former as a classification task and the latter as a sequence labeling task, this approach unifies both into a label generation problem (Fig. \ref{interaction} (d)). 

\citet{wu2022incorporating} proposed the first prompt-based Unified Generative Framework (UGEN), which identifies intents and slot labels in utterances through a question-answering formulation. Subsequently, \citet{ma2024promoting,wang2025unified} refined this framework by designing more fine-grained questions, thereby further exploring the potential of generative models for multi-intent SLU. Similar to UGEN, \citet{song2024unified} also reformulated multi-intent SLU as a question-answering task. \citet{xing2024dc} further enhanced the question-answering formulation of UGEN by explicitly integrating contrastive learning and bidirectional information transfer between tasks through prompting. Notably, this interactive paradigm eliminates the need for carefully designed information transfer modules. Instead, it leverages explicit knowledge from golden labels appended to the base prompts during training to guide the other task. Moreover, LLMs have demonstrated promising results in multi-intent SLU \cite{yin2024large}. \citet{yin2025eclm} incorporated chain-of-thought reasoning into multi-intent recognition, enabling deeper interpretability and achieving competitive performance.

\noindent\textbf{Advantages}: Directly generating intents and slot labels embedded in the utterance mitigates task discrepancies arising from conventional text classification and sequence labeling approaches, leading to a more unified modeling paradigm. This approach exhibits substantial potential for multi-intent SLU.

\noindent\textbf{Limitations}: The generative label prediction paradigm typically requires a generative PTM \cite{raffel2020exploring} or LLM \cite{he2024telechat,wang2025technical}, whereas traditional SLMs do not support such methods, resulting in higher computational and hardware demands.

Some of the projects discussed in $\S$ \ref{sec31} and $\S$ \ref{sec35} have been open-sourced. For ease of reference, we summarize the links to these publicly available projects in Table \ref{tab:link}.

\begin{table}[h]
\centering
\caption{Publicly available multi-intent SLU repositories.}
\label{tab:link}
\begin{tabular}{ll}
\toprule
\textbf{Model} & \textbf{Link} \\
\midrule
GISCo & \url{https://github.com/smxiao/GISCo} \\
PIAN & \url{https://github.com/puhahahahahaha/SLU_with_Co_PRoE} \\
Co-guiding Net & \url{https://github.com/XingBowen714/Co-guiding} \\
ReLa-Net & \url{https://github.com/XingBowen714/ReLa-Net} \\
MISCA & \url{https://github.com/VinAIResearch/MISCA} \\
MIVS\_BIRGAT & \url{https://github.com/importpandas/MIVS_BIRGAT} \\
BiSLU & \url{https://github.com/anhtunguyen98/BiSLU} \\
AGIF & \url{https://github.com/LooperXX/AGIF} \\
DGM & \url{https://github.com/dzy1011/DGM} \\
GL-GIN & \url{https://github.com/yizhen20133868/GL-GIN} \\
UGEN & \url{https://github.com/Young1993/UGEN} \\
ECLM & \url{https://github.com/SJY8460/ECLM} \\
EN-LLM & \url{https://github.com/SJY8460/SLM} \\
PromptSLU & \url{https://github.com/F2-Song/PromptSLU} \\
BERT-SIF & \url{https://github.com/heavenCSH/Few-Shot-SLU-BERT-SIF} \\
\bottomrule
\end{tabular}
\end{table}

\subsection{Comprehensive Analysis}\label{sec_res}
We report the performance of classification-based models and generation-based models on the MixATIS and MixSNIPS datasets in Tables~\ref{tab:res_cla} and \ref{tab:res_gen}, respectively. The main observations are summarized as follows:

\begin{table}[h]
\setlength{\tabcolsep}{0.4mm}
\renewcommand\arraystretch{1}
\centering
\caption{Experimental Results of Classification-based Models on the MixATIS and MixSNIPS.}
\label{tab:res_cla}
\begin{tabular}{lcccccc}
\toprule
\multirow{2}{*}{Models} & \multicolumn{3}{c}{MixATIS} & \multicolumn{3}{c}{MixSNIPS} \\
\cmidrule(lr){2-4} \cmidrule(lr){5-7}
 & Overall(Acc) & Slot(F1) & Intent(Acc) & Overall(Acc) & Slot(F1) & Intent(Acc) \\
\midrule
JMID-SF& 36.1 & 84.6 & 73.4 & 62.9 & 90.6 & 95.1 \\
AGIF& 40.8 & 86.7 & 74.4 & 74.2 & 94.2 & 95.1 \\
GL-GIN & 43.5 & 88.3 & 76.3 & 75.4 & 94.9 & 95.6 \\
MAG & 44.7 & 88.7 & 77.1 & 80.0 & 95.6 & 98.1\\
CLID & 49.0 & 88.2 & 77.5 & 75.0 & 94.3 & 96.6\\
ELSF & 49.1 & 88.7 & 75.9 & 75.3 & 94.7 & 96.1 \\
AGIF (RoBERTa-base) & 48.4 & 86.3 & 80.1 & 81.7 & 95.2 & 96.8 \\
GL-GIN (RoBERTa-base)& 50.1 & 86.9 & 80.8 & 82.6 & 96.4 & 97.3 \\
TFMN (BERT-base) & 50.2 & 88.0 & 79.8 & 84.7 & 96.4 & 97.7 \\
Uni-MIS (RoBERTa-base) & 52.5 & 88.3 & 78.5 & 83.4 & 96.4 & 97.2\\
\midrule
SLIM (BERT-base) & 47.6 & 88.5 & 78.3 & 84.0 & 96.5 & 97.2\\
\midrule
SDJN & 44.6 & 88.2 & 77.1 & 75.7 & 94.4 & 96.5 \\
IR-SF & 45.2 & 87.9 & 78.2 & 77.5 & 95.2 & 96.2 \\
CKEM & 46.6 & 89.7 & 77.1 & 79.7 & 96.7 & 96.1 \\
GISCo & 48.2 & 88.5 & 75.0 & 75.9 & 95.0 & 95.5 \\
DIF & 49.3 & 88.2 & 75.8 & 75.9 & 94.4 & 95.3 \\
GCN-MT & 51.4 & 89.5 & 75.6 & 76.4 & 95.1 & 97.0\\
ReLa-Net& 52.2 & 90.1 & 78.5 & 76.1 & 94.7 & 97.6 \\
Co-guiding& 51.3 & 89.8 & 79.1 & 77.5 & 95.1 & 97.7 \\
SSRAN & 48.9 & 89.4 & 77.9 & 77.5 & 95.8 & 98.4 \\
Co-guiding + TKDF  & 50.8 & 89.6 & 78.8 & 77.3 & 94.6 & 97.4\\
ReLa-Net + TKDF  & 51.2 & 89.8 & 78.4 & 75.9 & 94.2 & 97.0 \\
SSRAN + TKDF  & 58.5 & 89.2 & 77.6 & 77.2 & 95.4 & 98.1 \\
AGIF + Aligner$^2$  & 41.7 & 87.4 & 74.6 & 74.5 & 94.6 & 95.2\\
GLGIN + Aligner$^2$  & 44.2 & 88.7 & 76.6 & 76.0 & 95.3 & 95.9\\
SDGN + Aligner$^2$  & 45.1 & 88.7 & 77.4 & 76.2 & 94.8 & 96.9\\
GISCo + Aligner$^2$  & 49.0 & 89.1 & 75.7 & 76.5 & 95.6 & 95.8\\
Co-guiding + Aligner$^2$  & 51.9 & 90.4 & 79.6 & 77.9 & 95.6 & 98.0\\
ReLa-Net + Aligner$^2$  & 52.8 & 90.6 & 78.7 & 76.5 & 95.3 & 98.1\\
SSRAN + Aligner$^2$  & 49.4 & 89.7 & 77.9 & 77.8 & 96.0 & 98.6\\
BiSLU  & 51.5 & 89.4 & 81.5 & 85.4 & 97.2 & 97.8 \\
MISCA  & 53.0 & 90.5 & 76.7 & 77.9 & 95.2 & 97.3 \\
InfoJoint  & 51.9 & 90.6 & 79.8 & 78.2 & 96.9 & 97.9 \\
DHLG  & 54.9 & 92.3 & 81.5 & 79.1 & 96.4 & 98.2 \\
SDJN (BERT-base)  & 46.3 & 87.5 & 78.0 & 79.3 & 95.4 & 96.7\\
GISCo (RoBERTa-base) & 52.5 & 87.3 & 81.0 & 82.6 & 97.2 & 97.2 \\
ReLa-Net (RoBERTa-base) & 54.9 & 89.2 & 82.3 & 83.5 & 96.7 & 97.8 \\
Go-guiding (RoBERTa-base)& 54.3 & 88.4 & 83.2 & 83.9 & 97.6 & 98.1 \\
PIAN (BERT-base)  & 49.5 & 88.2 & 80.9 & 86.1 & 97.1 & 97.2 \\
DHLG (RoBERTa-base) & 56.7 & 90.4 & 84.6 & 86.4 & 97.6 & 98.8 \\
\bottomrule
\end{tabular}
\end{table}

\begin{table}[h]
\setlength{\tabcolsep}{0.4mm}
\renewcommand\arraystretch{1}
\centering
\caption{Experimental Results of Generation-based Models on the MixATIS and MixSNIPS.}
\label{tab:res_gen}
\begin{tabular}{lcccccc}
\toprule
\multirow{2}{*}{Models} & \multicolumn{3}{c}{MixATIS} & \multicolumn{3}{c}{MixSNIPS} \\
\cmidrule(lr){2-4} \cmidrule(lr){5-7}
 & Overall(Acc) & Slot(F1) & Intent(Acc) & Overall(Acc) & Slot(F1) & Intent(Acc) \\
\midrule
UGEN (T5-base) & 55.4 & 89.1 & 83.1 & 79.1 & 94.8 & 96.8 \\
UGEN (T5-large) & 57.9 & 89.6 & 84.2 & 81.0 & 95.8 & 97.2 \\
UGen-DP (T5-base) & 58.7 & 90.3 & 86.2 & 84.7 & 96.6 & 97.6 \\
PromptSLU (T5-base) & 57.2 & 89.6 & 85.8 & 84.8 & 96.5 & 97.5 \\
LSE-NLU (T5-base) & 58.2 & 90.1 & 85.3 & 84.5 & 96.3 & 97.9 \\
DC-Instruct (T5-base) & 58.1 & 90.4 & 84.4 & 81.2 & 95.7 & 97.6 \\
DC-Instruct (T5-large) & 60.5 & 90.7 & 84.9 & 83.9 & 96.4 & 97.8 \\
ChatGPT (gpt-3.5-turbo)  & 1.90 & 34.5 & 22.1 & 1.40 & 30.0 & 67.5 \\
EN (Vicuna-7B) & 47.3 & 83.3 & 79.5 & 78.9 & 95.7 & 97.6 \\
EN (LLaMA2-7B) & 51.1 & 86.5 & 82.4 & 78.9 & 95.7 & 96.9 \\
EN (Mistral-7B) & 53.4 & 88.7 & 80.6 & 79.8 & 95.6 & 97.6 \\
ENSI (Vicuna-7B) & 49.4 & 86.0 & 80.6 & 82.4 & 96.2 & 97.1 \\
ENSI (LLaMA2-7B) & 46.6 & 84.7 & 79.3 & 83.5 & 96.5 & 97.6 \\
ENSI (Mistral-7B) & 49.8 & 86.4 & 78.0 & 83.9 & 96.7 & 97.5 \\
ECLM (LLaMA3.1–8B) & 56.2 & 90.2 & 80.7 & 86.5 & 97.0 & 97.0 \\
UGEN (LLaMA2-7B) & 62.4 & 90.1 & 94.2 & 82.0 & 96.4 & 96.1 \\
UGEN (LLaMA2-13B) & 64.3 & 90.3 & 96.1 & 84.0 & 96.7 & 96.3 \\
DC-Instruct (LLaMA2-7B) & 64.1 & 90.7 & 94.6 & 84.0 & 96.7 & 96.3 \\
DC-Instruct (LLaMA2-13B) & 66.7 & 91.7 & 96.9 & 84.6 & 96.9 & 97.1 \\
\bottomrule
\end{tabular}
\end{table}

\begin{itemize}
\item \textbf{Impact of information flow direction.} Bidirectional interactive models exhibit greater potential for achieving superior performance compared with unidirectional guidance models. Their design objective is to establish effective information exchange between multiple intent detection and slot filling. Through interactive guidance, these models are able to capture utterance semantics and task correlations more comprehensively, thereby achieving higher accuracy.

\item \textbf{Impact of PTMs.} When employing the same model architecture, replacing the original encoder with a PTM (BERT \cite{liu2019roberta} or RoBERTa \cite{devlin2019bert}) generally leads to improvements in intent accuracy, slot F1, and overall accuracy. This gain arises because PTMs encode richer semantic knowledge than SLMs, making them a key driver of performance enhancement.

\item \textbf{Influence of base model and prompt design.} The performance of generative models largely depends on the underlying base model. Under the same prompting strategy, LLMs generally outperform smaller generative PTMs. For instance, the LLaMA2-7B-based \cite{touvron2023llama} UGEN model surpasses its T5-based \cite{raffel2020exploring} counterpart, and performance tends to improve further with larger model sizes — for example, UGEN built on LLaMA2-13B achieves better results than its 7B variant. Nevertheless, prompt design remains a crucial factor affecting performance, as well-crafted prompts can partially offset model size limitations. For example, models such as UGEN and UGEN-DP built on T5 outperform several LLM-based counterparts (e.g., EN and ENSI built on Mistral-7B \cite{jiang2024mixtral}, Vicuna-7B \cite{chiang2023vicuna}, or LLaMA3.1–8B \cite{dubey2024llama}) across all three evaluation metrics.

\item \textbf{Comparative potential of modeling paradigms.} From a holistic perspective, generative models demonstrate greater potential. This advantage may stem from their unified treatment of the two tasks and the strong representational capacity of LLMs, which jointly enable more accurate intent and slot recognition. Overall, both classification-based and generation-based models have their respective merits: classification-based models are more computationally efficient, whereas generation-based models exhibit higher potential for achieving superior accuracy.

\end{itemize}

\section{Future Challenges and Research Prospects}\label{sec4}
In this section, we present a discussion of the current challenges and future research directions for multi-intent SLU, focusing on three key aspects: datasets ($\S$ \ref{sec41}), modeling approaches and performance ($\S$ \ref{sec42}), and practical applications ($\S$ \ref{sec43}).

\subsection{Datasets}\label{sec41}
Since research on multi-intent SLU emerged later than that on single-intent SLU, the field still lacks large-scale and diverse datasets, particularly those spanning multiple domains and languages.

\subsubsection{Real-world Multi-intent SLU Datasets}\label{sec411}
Existing multi-intent SLU studies primarily rely on the MixATIS and MixSNIPS datasets, both of which are extended from single-intent SLU datasets rather than collected from real multi-intent scenarios. \citet{casanueva2022nlu++} introduced NLU++, the first multi-intent SLU dataset derived from real-world interactions. However, real-world scenarios are inherently diverse and complex, while current datasets mainly capture parallel relationships among intents without modeling potential inter-intent dependencies such as sequential or hierarchical relations \cite{xu2024birgat}. Therefore, expanding multi-intent SLU research to datasets that better reflect real-world applications across diverse domains is a pressing need. Such strategies, including incorporating the user’s surrounding environment and current state \cite{xu2022text,wu2025introducingvisualscenesreasoning} or integrating relevant knowledge bases, are analogous to approaches used in question answering \cite{zhang2023ccl23,song2015feature,song2022beyond}.

\subsubsection{Multi-lingual Multi-intent SLU Datasets}\label{sec412}
There are over 6,000 languages worldwide \cite{leben2018languages}, and researchers have made continuous efforts to develop single-intent SLU datasets across various languages. For instance, the CAIS dataset is designed for Chinese SLU \cite{liu2019cm}, Multi-ATIS++ covers nine languages \cite{xu2020end}, and MTOP includes six languages \cite{li2021mtop}. Furthermore, \citet{fitzgerald2023massive} extended SLURP \cite{bastianelli2020slurp} to 51 languages, releasing the MASSIVE dataset, which was later expanded to include Uyghur \cite{aimaiti2024uyghur}.

In contrast, progress in multi-intent SLU under multilingual settings has been relatively limited. \citet{moghe2023multi3nlu++} introduced MULTI$^3$NLU++, an extension of NLU++ that incorporates four additional languages, including Spanish, Marathi, Turkish, and Amharic. Compared with single-intent SLU, multilingual datasets for multi-intent SLU remain scarce and cover fewer languages. Therefore, advancing research on multi-intent SLU to support diverse linguistic scenarios remains both meaningful and necessary.

\subsection{Modeling Approaches and Performance}\label{sec42}
In multi-intent SLU, an utterance may contain multiple intent clauses, which makes the task inherently more challenging than single-intent SLU. Achieving strong performance thus requires more sophisticated modeling approaches capable of capturing complex intent interactions and contextual dependencies. Enhancing the robustness and generalization ability of multi-intent SLU models across diverse scenarios remains an important direction for future research.

\subsubsection{More Competitive Performance}\label{sec421}
The performance comparison of the respective task SOTA models for multi-intent SLU and single-intent SLU in supervised settings is shown in Table~\ref{tab:result_com}.
\begin{table}[h]
\renewcommand\arraystretch{1}
\centering
\caption{Performance of the SOTA models for single-intent and multi-intent SLU. Among them, the results on ATIS and SNIPS are obtained from HAN \cite{chen2022towards}, while the results on MixATIS and MixSNIPS are derived from DC-Instruct \cite{xing2024dc}.}
\label{tab:result_com}
\begin{tabular}{lcccc}
\toprule
Metric & ATIS & MixATIS & SNIPS & MixSNIPS\\
\midrule
Intent (Acc) & 98.5 & 96.9 & 99.2 & 97.1\\
Slot (F1) & 96.8 & 91.7 & 97.7 & 96.9\\
Overall (Acc) & 89.3 & 66.7 & 93.5 & 84.6\\
\bottomrule
\end{tabular}
\end{table}

As shown, MixATIS and MixSNIPS are derived from the single-intent ATIS and SNIPS datasets, respectively. However, the intent accuracy, slot F1, and overall accuracy achieved on multi-intent SLU datasets still lag behind those of SOTA models on single-intent datasets. 
Enhancing the interaction between multiple intent detection and slot filling or refining the decoding strategies for intents and slots may offer promising directions for improving model performance.

Enhancing the interaction between multiple intent detection and slot filling, optimizing the decoding strategies for intent and slot representations, introducing beneficial positive noise for regularization or stimulation \cite{PN,VPN,PiNDA,PiNI,PiNGDA,MiN,RN,MuNG,Laytrol}, or adopting higher-performing PTMs or LLMs are all potential and effective approaches for improving overall model performance \cite{li2024tele,wang2024telechat,li202452b}.

\subsubsection{Learning with limited training data}\label{sec422}
In real-world scenarios, obtaining sufficient supervised training data remains a major challenge. While single-intent SLU has achieved remarkable progress in few-shot learning \cite{jiang2025utterance} through models such as Retriever \cite{yu2021few}, EJSC \cite{liu2021explicit}, ConProm \cite{hou2021learning}, ILLUMINER \cite{mirza2024illuminer}, and JMRM \cite{han2024decoupling}, the advancement of multi-intent SLU has been relatively limited. Although several studies have evaluated their proposed methods under few-shot settings \cite{xing2024dc}, most efforts primarily emphasize improving multiple intent detection rather than addressing the full SLU pipeline \cite{wu2021label,hou2021few,geng2024large,zhou2024two,qin2025divide,zhuang2025prompt}. Enhancing the performance of multi-intent SLU in zero-shot and few-shot scenarios is therefore crucial, as it can mitigate the difficulties associated with constructing large-scale supervised datasets \cite{hua2024unraveling}. Furthermore, adopting weakly supervised or semi-supervised learning paradigms could further reduce annotation costs while maintaining acceptable model performance.

\subsubsection{Interpretability of multi-intent SLU}\label{sec423}
The interpretability of deep learning models has long been a central research focus, and this is equally important for multi-intent SLU. \citet{zhuang2024towards} examined model interpretability from the perspective of entropy in intent prediction, whereas \citet{yin2025eclm} employed a chain-of-thought approach to segment utterances into multiple intent clauses and identify the intent of each clause individually, providing a certain level of interpretability. However, this method is primarily applicable to utterances containing the conjunction ``and”. \citet{xing2022group,wu2024dual} visualized utterance embeddings using t-SNE \cite{maaten2008visualizing}, yet such representations are not readily interpretable for end users. Therefore, advancing research on the interpretability of multi-intent SLU models remains an important direction.

\subsection{Practical Applications}\label{sec43}
In practical applications, multi-intent SLU models must account not only for accuracy but also for other critical factors, including integration with other components of task-oriented dialogue systems, model size and inference efficiency, and the ability to handle unseen labels and support continual learning.

\subsubsection{Joint Learning with Downstream Dialogue Components}\label{sec431}
The SLU module is a critical component of task-oriented dialogue systems. In traditional pipeline-based systems, it is trained as an independent module \cite{song2017intension,deriu2021survey}. While this approach provides a certain level of controllability, it is susceptible to error propagation across the pipeline \cite{qin2023end}. A more recent trend involves the joint training of multiple modules \cite{su2024domain,su2025raicl}, which preserves modular controllability while reducing error accumulation \cite{he2022unified,mo2024hiertod}. Jointly learning multi-intent SLU with other system components \cite{song2024improving,song2024graph,su2025raicl,xie2025mitigating,xie2022correctable,su2024domain,su2023scalable} for end-to-end task-oriented dialogue \cite{king2024unsupervised} presents a promising avenue for enhancing overall response capability. Additionally, incorporating other dialogue-related tasks can also be a meaningful direction, such as emotion dialogue \cite{liu2024speak,yao2020session} or empathetic dialogue \cite{jiang2023empathy,jiang2023emp,jiang2025label,jiang2025utterance,zhao2023muse,zhao2024autograph,wang2024towards,wang2025emotional,song2021emotion,zhao2025mege,wang2025empathetic}. While assisting users in accomplishing specific tasks, model responses should also attend to users’ emotional needs as much as possible.

\subsubsection{Model Size and Inference Efficiency}\label{sec432}
When deploying multi-intent SLU models in real-world scenarios, model size and inference efficiency are critical considerations. Some models are built upon LLMs \cite{yin2024large,xing2024dc}, which achieve strong performance; however, their substantial parameter sizes pose challenges for deployment on downstream devices. Therefore, model size should be reported alongside accuracy when evaluating performance. In addition, inference efficiency directly impacts user experience. For example, \citet{qin2021gl} employed a non-autoregressive decoding strategy for intent and slot prediction, which has since been widely adopted \cite{song2022enhancing,wu2024dual}. In summary, the development of multi-intent SLU models should balance improvements in accuracy with hardware constraints and inference costs to ensure practical applicability.

\subsubsection{Continual Learning for Newly Emerging Classes}\label{sec433}
Most multi-intent SLU models are trained on a predefined set of candidate labels. In real-world applications, however, the set of labels is not fixed, and changes in the label set can make retraining the entire model costly. Continual learning provides an effective approach to accommodate the expansion of label sets \cite{wang2024comprehensive,mao2025classifier}. For example, \citet{song2023continual} applied continual learning to intent detection, while \citet{hui2021joint} extended it to full single-intent SLU. Progress in continual learning for multi-intent SLU remains limited, and further research in this area would support the practical deployment of multi-intent SLU systems.

\section{Conclusion}\label{sec5}
This paper presents a comprehensive review of research on multi-intent SLU. We categorize and summarize existing studies according to their decoding strategies and modeling paradigms, analyzing their respective advantages and drawbacks. Moreover, we discuss the current limitations in multi-intent SLU research and highlight promising directions, challenges, and opportunities for future exploration. To support further research, we also compile available multi-intent SLU datasets and open-source baseline projects. We hope this survey can serve as a valuable reference and foster continued progress in the field of multi-intent SLU.



\section*{Abbreviation}
\begin{tabularx}{\textwidth}{lX}
SLU & Spoken Language Understanding \\
PTM & Pre-trained Model \\
CRF & Conditional Random Fields \\
SVM & Support Vector Machines \\
SLM & Small Language Model \\
LLM & Large Language Model \\
AGIF & Adaptive Graph-Interactive Framework \\
DGM & Dynamic Graph Model \\
GLGIN & Global-Locally Graph Interaction Network \\
CLID & Chunk Level Intent Detection \\
SLIM & Slot-Intent Mapping Model \\
SDJN & Self-Distillation Joint Neural Language Understanding Model \\
CKEM & Conceptual Knowledge Enhanced Model \\
GCN-MT & Graph Convolutional Network and Multi-head Transformer \\
SSRAN & Scope-Sensitive Result Attention Network \\
ReLa-Net & Recurrent Heterogeneous Label Matching Network \\
UGEN & Unified Generative Framework \\
\end{tabularx}


\section*{Declarations}

\subsection*{Authors' contributions}
\textbf{Di Wu}: Writing – original draft, collection and summary of references.
\textbf{Ruiyu Fang}: Writing – original draft, collection and summary of references.
\textbf{Liting Jiang}: Writing – original draft, collection and summary of references.
\textbf{Shuangyong Song}: Writing – original draft, collection and summary of references.
\textbf{Xiaomeng Huang}: Image.
\textbf{Shiquan Wang}: Image.
\textbf{Zhongqiu Li}: Decoding method summary.
\textbf{Lingling Shi}: Decoding method summary.
\textbf{Mengjiao Bao}: Summary of the Multi-Intent SLU Model.
\textbf{Yongxiang Li}: Financial support.
\textbf{Hao Huang}: Financial support.

\subsection*{Competing Interest}
The authors declare that they have no known competing financial interests or personal relationships that could have appeared to influence the work reported in this paper.

\subsection*{Acknowledgements}
Not applicable.

\subsection*{Data availability}
Not applicable.

\subsection*{Code availability}
Not applicable.

\subsection*{Funding }
Not applicable.

\bibliography{sn-bibliography}

\end{document}